\documentclass{article}
\usepackage{spconf,amsmath,epsfig}
\usepackage{url}
\usepackage{graphicx}
\usepackage{subcaption}
\usepackage{ifthen}
\usepackage{float}

\begin{document}\sloppy
\def\x{{\mathbf x}}
\def\L{{\cal L}}

\title{Panoptic-Depth Color Map for Combination of Depth and image segmentation}
\name{$^1$Jia-Quan Yu, $^2$Soo-Chang Pei}
\address{$^1$Graduate Institute of Communication Engineering,\\
National Taiwan University, Taiwan\\
E-mail: kenyu910645@gmail.com\\
$^2$Department of Electrical Engineering,\\
National Taiwan University, Taiwan\\
E-mail: peisc@ntu.edu.tw}
\maketitle

\begin{abstract}
Image segmentation and depth estimation are crucial tasks in computer vision, especially in autonomous driving scenarios. Although these tasks are typically addressed separately, we propose an innovative approach to combine them in our novel deep learning network, Panoptic-DepthLab. By incorporating an additional depth estimation branch into the segmentation network, it can predict the depth of each instance segment. Evaluating on Cityscape dataset, we demonstrate the effectiveness of our method in achieving high-quality segmentation results with depth and visualize it with a color map. Our proposed method demonstrates a new possibility of combining different tasks and networks to generate a more comprehensive image recognition result to facilitate the safety of autonomous driving vehicles.
\end{abstract}


\section{Introduction}
\label{sec:intro}

Image segmentation is a crucial task in Computer Vision that involves partitioning camera images into different segments or instances based on the semantic meaning of each pixel. It has a wide range of applications in fields such as medical image processing, image processing, and autonomous vehicles. There are three types of segmentation tasks: semantic, instance, and panoptic segmentation. Semantic segmentation is equivalent to pixel classification while each pixel belongs to a semantic category, while instance segmentation focuses on segmenting out the foreground objects and differentiating different instances within the scene. The third task, panoptic segmentation, is a relatively new task that aims to unify semantic and instance segmentation into a single task by performing semantic segmentation on background pixels and instance segmentation on foreground pixels. 

\begin{figure}
\centering
\includegraphics[width=8.5cm]{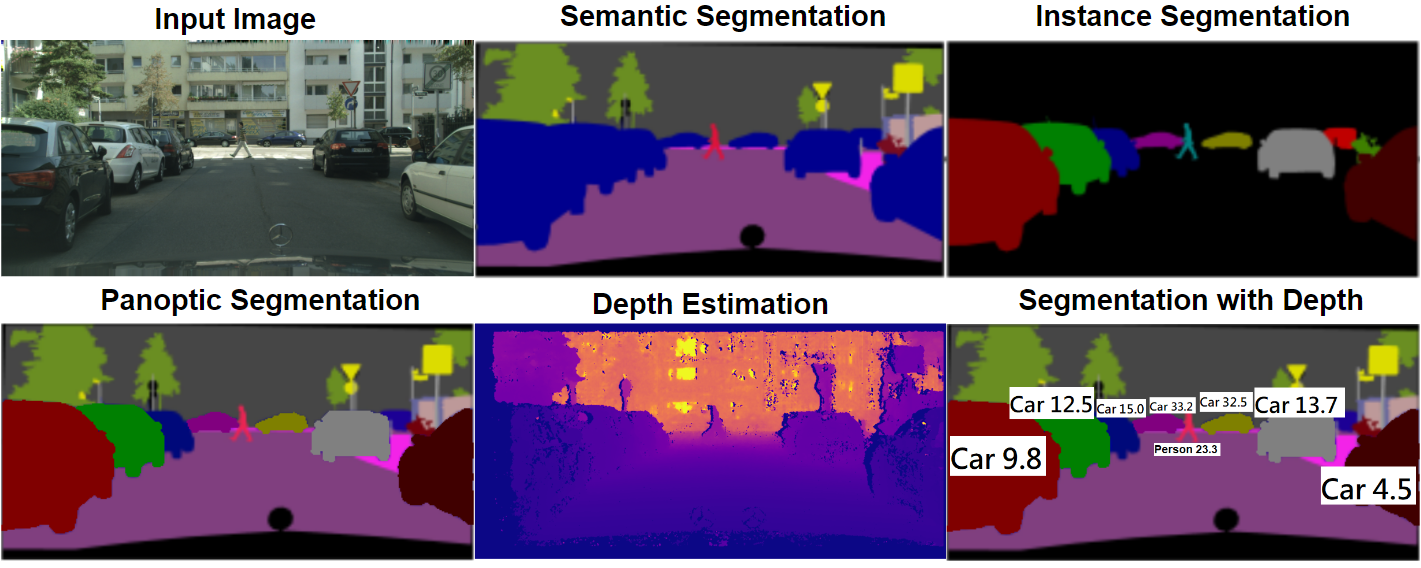}
\caption{Image task taxonomy. Semantic segmentation classifies each image pixel into semantic categories. Instance segmentation identifies different instances within the foreground pixels. Panoptic segmentation combines both instance and semantic segmentation tasks. Depth estimation evaluates the depth value of each pixel. Lastly, our proposed segmentation with depth task aims to integrate depth estimation and panoptic segmentation, producing a color map that represents the instance-level depth value.}
\label{fig:taxonomy}
\end{figure}

Another important task in autonomous driving scenarios is depth estimation. It involves predicting the depth value of each image pixel, where the depth represents the distance from the camera center to the nearest obstacle. Depth estimation is particularly essential in autonomous driving as it provides information for the navigation system to avoid collisions. However, with only pixel-wise depth information, it is insufficient to distinguish instances in the scene. Therefore, we propose to combine image segmentation and depth estimation  to predict depth value on the instance level and generate a color map to visualize the image recognition result. We depict the image tasks taxonomy we introduced in this section in Fig.\ref{fig:taxonomy}.

\begin{figure*}
\centering
\includegraphics[width=17cm]{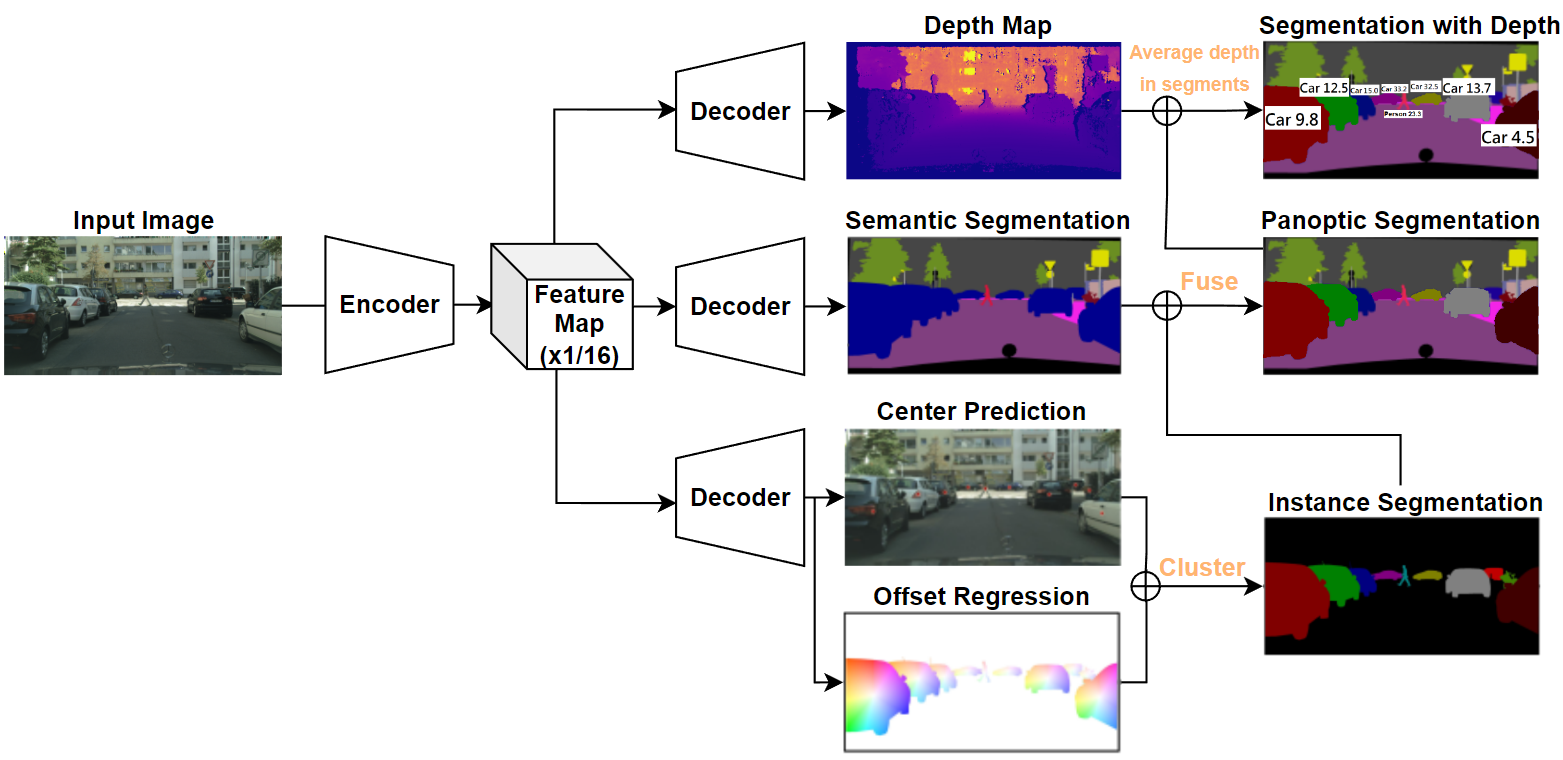}
\caption{Panoptic-DepthLab architecture. Based on the Panoptic-DeepLab\cite{pdeeplab} framework, we extend it by adding a depth estimation decoder branch to predict depth map. The second decoder branch is responsible for predicting semantic segmentation result, as the third branch predict the instance segmentation outcome. To incorporate depth and segmentation results, we average out the region of each correspondent segment in a depth map to acquire the instance-level depth value.}
\label{fig:architecture}
\end{figure*}

To achieve our proposed task, we introduce Panoptic-DepthLab, an extended variant of the Panoptic-DeepLab\cite{pdeeplab} segmentation network. By incorporating an additional depth estimation branch into the panoptic segmentation network, it is able to generate segmentation and depth estimation results at the same time. Subsequently, we fuse the panoptic segmentation and depth estimation results to generate a color map that represents the depth of each instance. This color map can provide valuable insights for ensuring safety in autonomous driving scenarios. We illustrate our proposed Panoptic-DepthLab architecture in Fig.\ref{fig:architecture}.

The rest of the paper is organized as follows. Section 2 presents a related work review in panoptic segmentation and depth estimation. In section 3, we describe our proposed method, Panoptic-DepthLab in details. Section 4 reports our experimental results on the Cityscape dataset. Finally, section 5 concludes the paper.

\section{Related Works}

\subsection{Panoptic Segmentation}

Panoptic segmentation\cite{panoptic_segmentation}, also known as scene parsing, combines the tasks of semantic and instance segmentation. It involves classifying background pixels into semantic categories and classifying foreground pixels into individual instances. Many related works extend well-established instance segmentation networks by incorporating an additional sementic segmentation branch to achieve panoptic segmentation. For instance, Panoptic-FPN \cite{pfpn}, which is based on Mask-RCNN\cite{mask_rcnn} network, add a segmentation branch to handle background pixel categories. Additionally, it utilizes a Feature Pyramid Network (FPN) to extract detailed features for the segmentation branches.

DeeperLab\cite{deeperlab}, for other example, adopts the encoder-decoder architecture and Atrous Spatial Pyramid Pooling(ASPP) of DeepLab\cite{deeplabv3+} and adds an additional prediction head to predict semantic segments and four other keypoint-based detections to find instances. The UPSNet\cite{upsnet} architecture is similar to Panoptic-FPN, which adds an FPN module to the backbone and designs separate semantic and instance heads for prediction. UPSNet is able to achieves 61.8\% PQ on the Cityscape validation dataset.

Panoptic-DeepLab\cite{pdeeplab}, the network on which our work is based, extends the DeepLabv3+ network architecture to enable panoptic segmentation. It incorporates dual-ASPP and dual-decoder modules to produce three outputs: semantic segmentation, object center prediction, and center offset prediction. For the instance segmentation branch, Panoptic-DeepLab combines the center prediction heatmap with pixel offset regression to generate class-agnostic instances. It then employs a majority-vote algorithm, fusing the instance predictions with the semantic segmentation result to determine the instance categories. Panoptic-DeepLab adopts a bottom-up approach to address the instance segmentation aspect of the task, achieving a PQ (Panoptic Quality) score of 64.1\% on the Cityscape validation set.

In conclusion, these works achieve panoptic segmentation by extending instance segmentation networks to generate semantic segmentation results for background pixels. Among these works, Panoptic-DeepLab has demonstrated outstanding performance. Therefore, we have chosen it to be the basis network for our own approach, leveraging its superior performance.

\subsection{Monocular Depth Estimation}
Monocular depth estimation is another fundamental task in Computer Vision that aims to predict the distance of each image pixel from the camera center to the nearest obstacle using a single camera image as input. Traditional approaches adopt an encoder-decoder architecture similar to semantic segmentation networks design. The other line of work intent to model the ordinal relationship of the depth value. Deep Ordinal Regression Network (DORN) \cite{dorn} offers an effective approach by formulating the depth regression problem as an ordinal regression problem. DORN introduces multiple binary classification subtasks to infer the depth, where each subtask determines whether the pixel depth is greater than a specific threshold. The final depth value is obtained by summing the number of subtasks that predict the pixel to be deeper than each threshold. Additionally, DORN incorporates the ASPP module to capture global contextual information and enhance the decoder's performance. Given the effectiveness of DORN, we adopt this design and integrate it into our depth estimation branch.
\section{Proposed Method}

\subsection{Panoptic-DepthLab}
\begin{figure*}[hp!]
\centering
\includegraphics[width=17cm]{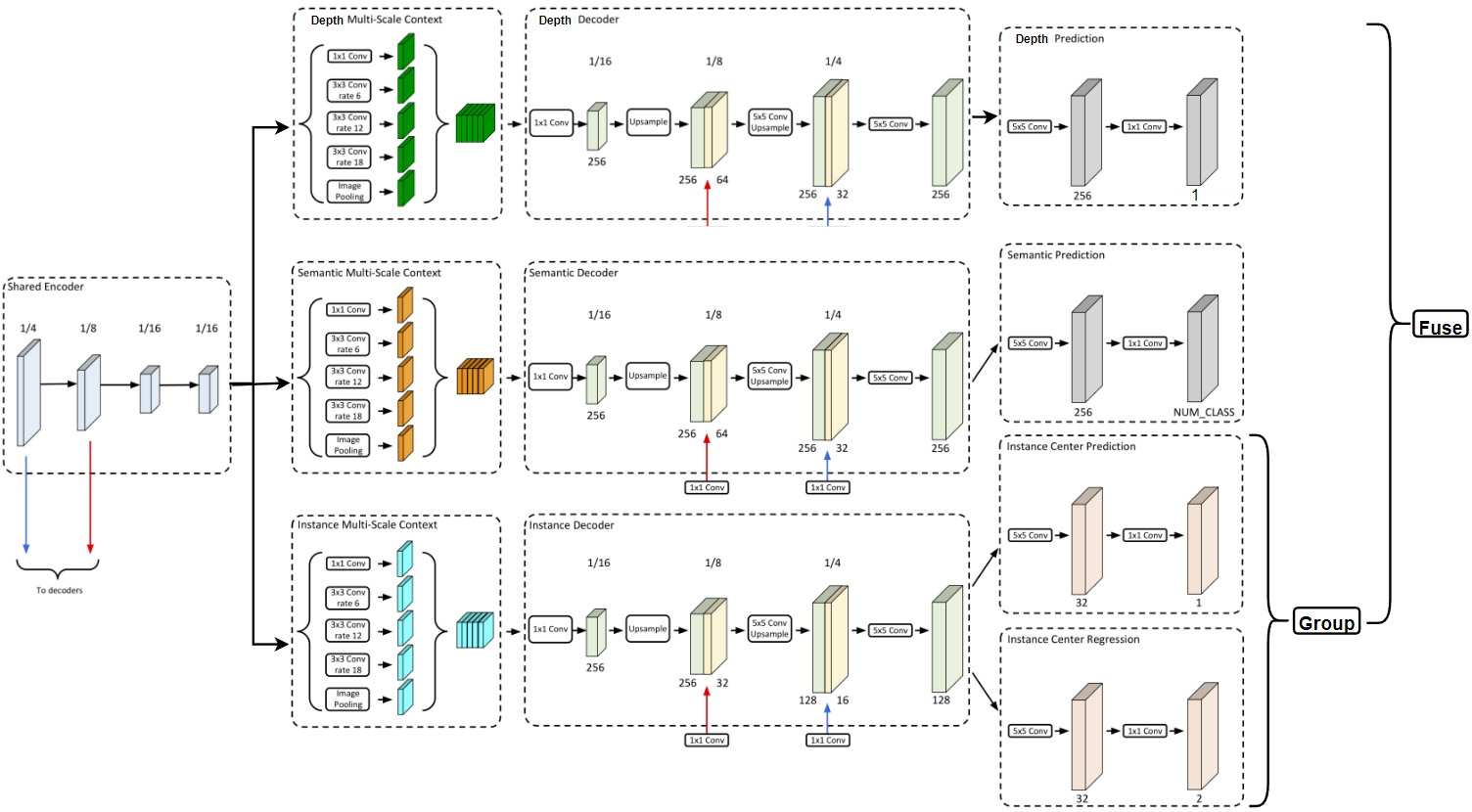}
\caption{Architecture of Panoptic-DepthLab. Based on Panoptic-DeepLab, we introduce an additional decoder branch specifically designed for depth estimation. The network consists of three branches: semantic segmentation, instance segmentation, and depth estimation, all utilizing the shared feature map extracted by the encoder. This unified network allows for end-to-end training with only single camera image as input. During inference, a post-processing step combines the segmented regions with depth information.}
\label{fig:detail_architecture}
\end{figure*}

In this section, we introduce Panoptic-DepthLab, our proposed network that aims to produce segmentation with instance-level depth. By adding a depth estimation decoder branch to the Panoptic-DeepLab segmentation network, we can get a unified one-stage network that effectively predict segmentation and depth map in parallel branches. Additionally, we let the segmentation and depth branch share the same extracted feature map as input. This design attempt to increase efficiency and avoid extracting duplicate feature map. During training, we optimize the whole model, allowing the network to jointly learn multiple tasks. To combine the predicted depth map with the panoptic segmentation result, we average the pixel depths within each instance region. The shared encoder features facilitate efficient end-to-end training. 
Specifically, our newly added depth estimation branch follows a similar design to the semantic segmentation branch. The detailed architecture is illustrated in Fig \ref{fig:detail_architecture}.

\section{Experiments}

In this section, we present the detail experimental setting and results of our Panoptic-DepthLab network, which is built upon the Detectron2\cite{detectron2} platform. We conduct our experiment on Cityscape\cite{cityscapes} dataset, which includes 2975 images for training and 500 images for validation, where most of the image is shot on cities street in Germany. The dataset consists of eight background categories and eleven foreground categories for the panoptic segmentation challenge.

During training, we initialize our network with the pre-trained weights provided by Panoptic-DeepLab, which were trained on the Cityscape dataset for 90K iterations. Subsequently, we fine-tune the entire network for an additional 10K iterations after incorporating our additional depth estimation branch. The training process is conducted on 2 TITAN RTX GPUs, with a batch size of 14. We employ the Adam optimizer with a learning rate set to 0.001. Additionally, since Cityscape does not provide ground truth depth estimation, we generate depth maps by converting the provided disparity map to depth map.

\subsection{Evaluation Metric}
To assess the quality of panoptic segmentation results, Kirillov et al. introduced a novel metric called Panoptic Quality (PQ) \cite{panoptic_segmentation}. PQ evaluates the accuracy of both semantic classification and masks prediction. The PQ for a class is defined as follows:

\begin{equation}
PQ=\overbrace{\frac{\sum_{(p,g)TP}IoU(p,g)}{|TP|}}^{\text{Segmentation Quality(SQ)}}\times\overbrace{\frac{|TP|}{|TP|+\frac{1}{2}|FP|+\frac{1}{2}|FN|}}^{\text{Recognition Quality(RQ)}}
\label{eq:PQ}
\end{equation}
where $|TP|$, $|FP|$, and $|FN|$ represent the number of true positive, false positive, and false negative respectively. 
The first term of the equation \ref{eq:PQ} is segmentation quality (SQ), which is the average IoU(Intersection over Union) of true positives masks. The second term is recognition quality (RQ), which is the F1 score of classification result.

To evaluate the depth estimation result, we adopt the following metrics:\\
Relative squared error: 
\begin{equation}
\text{sqErr} = \frac{1}{N}\sum_{i=1}^{N}(\frac{d^*_i - d_i}{d_i})^2
\end{equation}
where $d_i$ and $d_i^*$ represent ground truth depth value and predicted depth value. And the $N$ represent the number of pixel in image.

Relative absolute error:
\begin{equation}
\text{absErr} = \frac{1}{N}\sum_{i=1}^{N}|\frac{d^*_i - d_i}{d_i}|
\end{equation}

Inverse root mean square error:
\begin{equation}
\text{IRMSE} = \sqrt{\frac{1}{N}\sum_{i=1}^{N}\left(\frac{1}{d^*_i}-\frac{1}{d_i}\right)^2}
\end{equation}

Scale invariant logarithmic error:
\begin{equation}
\text{SILog} = \frac{1}{N}\sum_{i=1}^{N}x_i^2 - \frac{1}{N^2}(\sum_{i=1}^{N}x_i)^2
\end{equation}
where $x_i=logd_i - logd^*_i$

Accuracy with threshold:
\begin{equation}
\delta_i=max(\frac{d_i}{d_i^*}, \frac{d_i^*}{d_i}) < t
\end{equation}
where $t \in [1.25, 1.25^2, 1.25^3]$, It represents the percentage of image pixels whose depth value ratio with the ground truth is smaller than the given threshold $t$.

\subsection{Quantitative Result}
We present the Panoptic Quality (PQ) of the Panoptic-DepthLab model tested on the Cityscape validation dataset in Table \ref{tab:PQ_exp}. Our results show that Panoptic-DepthLab achieves slightly better performance compared to other panoptic segmentation networks. This improvement can be attributed to the additional depth map data available during our fine-tuning training, implicitly giving more useful information to the segmentation branch.

\begin{table}
  \centering
  \caption{Experimental results of Panoptic-DepthLab on the Cityscape validation set. The metric used is Panoptic Quality (PQ). Semantic Quality (SQ) represents the quality of predicted masks, while Recognition Quality (RQ) represents the classification accuracy. The PQ is calculated as $PQ = SQ \times RQ$. Panoptic-DepthLab performs slightly better than other network on PQ metirc.}
  \label{tab:PQ_exp}
  \begin{tabular}{c|cccc}
    Methods & PQ & SQ & RQ \\
    \hline
    Panoptic-FPN\cite{pfpn} & 55.4 & 77.9 & 69.3 \\
    UPSNet\cite{upsnet}     & 60.1 & 80.3 & \textbf{73.5} \\
    Panoptic-DepthLab       & \textbf{60.3} & \textbf{81.5} & 72.9 \\
  \end{tabular}
\end{table}

To determine the better loss function for our depth estimation branch, we compared two different loss functions. The first one is the loss function proposed by DORN\cite{dorn}, which discretizes depth values into intervals based on uncertainty and formulates the regression problem as multiple binary classification subtasks. The second loss function is smoothed L1 loss.

The results of our experiments are presented in Table \ref{tab:rse} and Table \ref{tab:delta}. Surprisingly, we found that smooth L1 loss outperformed DORN's loss after training for 10K iterations. This may be attributed to the complexity of the DORN method, which involves a more complex network structure and a larger number of additional parameters. In contrast, the smoothed L1 loss proved to be a more effective and straightforward method for our specific task.

\begin{table}
\centering
\caption{Evaluation of depth estimation results produced by Panoptic-DepthLab. Both loss function settings were trained for 10K iterations and tested on the Cityscape validation dataset. Evaluation metrics include relative squared error (sqErrorREL), relative absolute error (absErrorRel), inverse root mean square error (IRMSE), and scale invariant logarithmic error (SILog), which assess the difference between the predicted depth map and the ground truth depth map. Smaller values indicate better performance.}
\label{tab:rse}
\begin{tabular}{c|cccc}
Loss & sqErr & absErr & IRMSE & SILog \\
\hline
DORN\cite{dorn} & 0.72 & 0.69 & 44.64 & 22.31 \\
L1 & \textbf{0.63} & \textbf{0.60} & \textbf{34.57} & \textbf{18.58} \\
\end{tabular}
\end{table}

\begin{table}
\centering
\caption{Evaluation of depth estimation results produced by Panoptic-DepthLab. Both loss function settings were trained for 10K iterations and tested on the Cityscape validation dataset. The evaluation metric is accuracy with threshold, representing the percentage of depth map pixels whose ratio with the ground truth pixels is smaller than the assigned threshold value. Threshold values of 1.25, $1.25^2$, and $1.25^3$ are used. Higher values indicate better performance.}
\label{tab:delta}
\begin{tabular}{c|ccc}
Loss & $\delta_1<1.25$ & $\delta_2<1.25^2$ & $\delta_3<1.25^3$ \\
\hline
DORN & 0.27 & 0.49 & 0.68 \\
L1 & \textbf{0.30} & \textbf{0.61} & \textbf{0.82} \\
\end{tabular}
\end{table}

\subsection{Qualitative Result}

\begin{figure*}[hp!]
  \centering
  \includegraphics[width=0.9\textwidth]{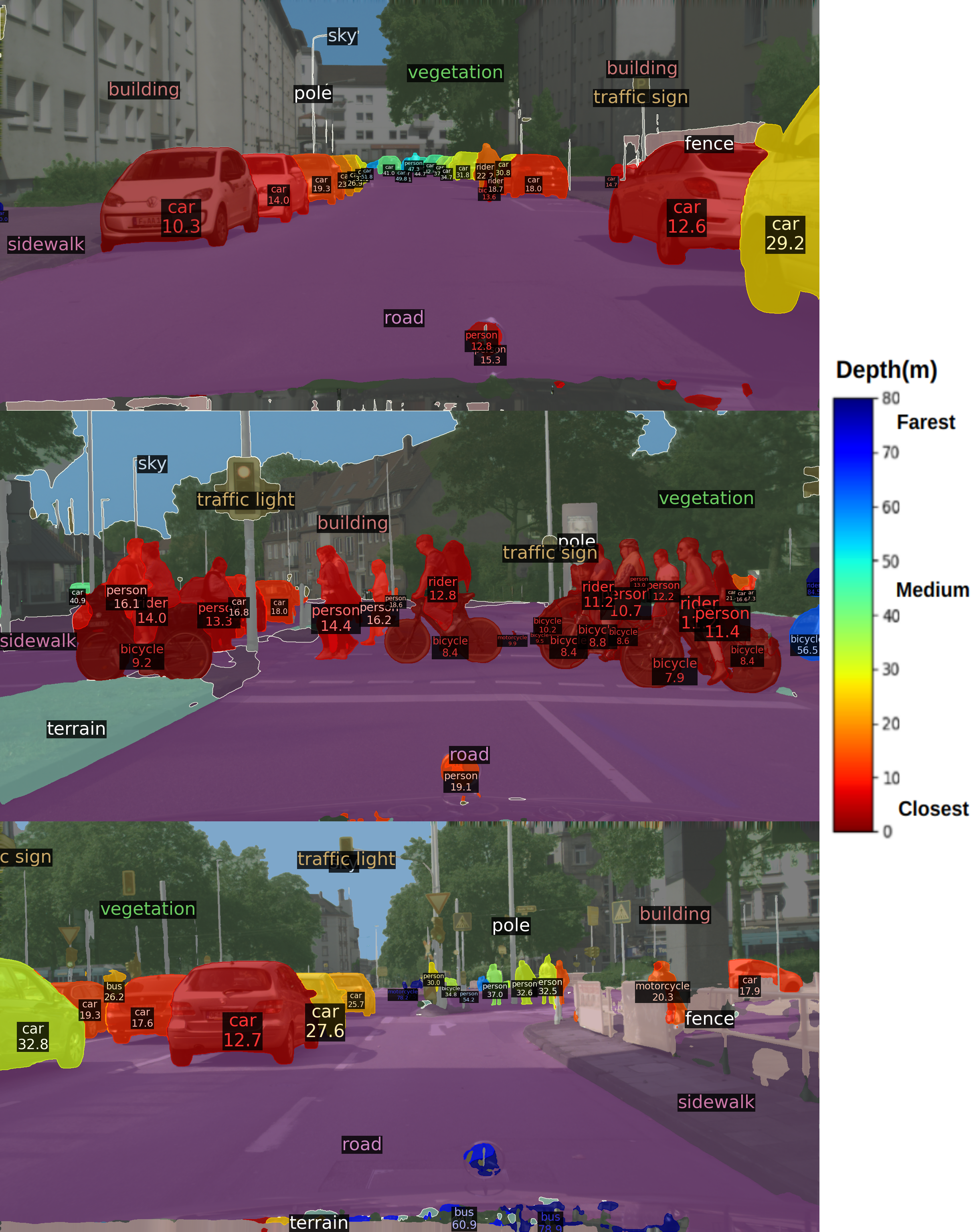}
  \caption{Inference example of Panoptic-DepthLab on the Cityscape validation set. Each instance is labeled with a category and a depth value, which determine the color of each foreground segment. The objects closest to the camera are colored in red, while the farthest objects are colored in blue. The background pixel is also segmented by its semantic meaning. Our inference example shows our network is able to produce high-quality segmentation with depth.}
  \label{fig:example_1}
\end{figure*}

\begin{figure*}[hp!]
  \centering
  \includegraphics[width=0.9\textwidth]{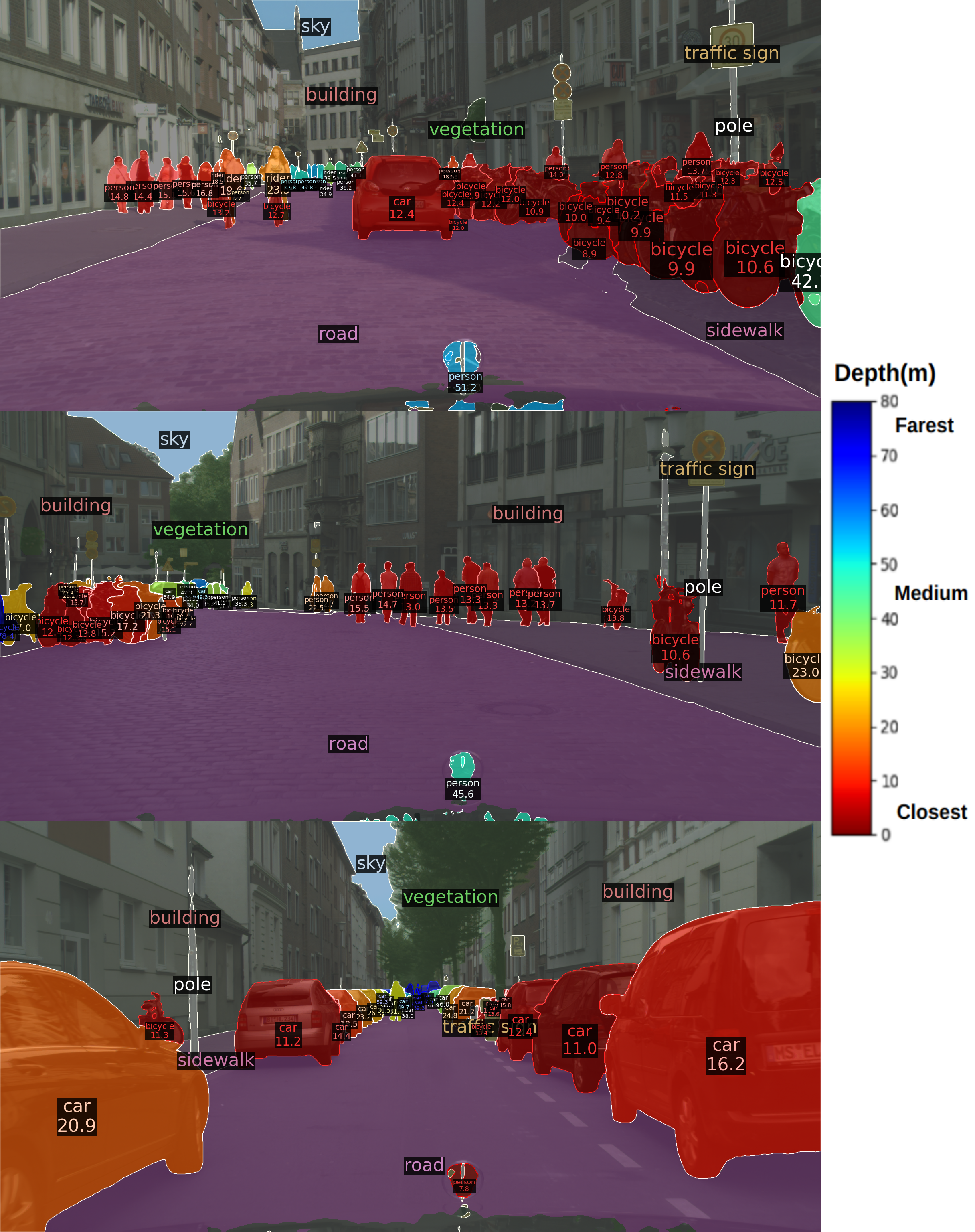}
  \caption{Inference example of Panoptic-DepthLab on the Cityscape validation set. Each instance is labeled with a category and a depth value, which determine the color of each foreground segment. The objects closest to the camera are colored in red, while the farthest objects are colored in blue. The background pixel is also segmented by its semantic meaning. Our inference example shows our network is able to produce high-quality segmentation with depth.}
  \label{fig:example_2}
\end{figure*}

To assess the performance of Panoptic-DepthLab, we visualize inference example on the Cityscape validation set, presented in Fig \ref{fig:example_1} and Fig \ref{fig:example_2}. In these examples, each predicted instance is visually distinguished by its assigned color based on their depth value. Objects in close proximity appear in vibrant red, while those with medium distance is colored in green and farther away objects are colored in cooler shades of blue. Additionally, the background pixels, such as road, sky, and vegetation, are effectively segmented and labeled in the resulting output. Each instance is also annotated with its respective category and depth information, providing a detail understanding of the scene.

\section{Conclusions}

In this paper, we have presented a novel approach to combine depth estimation and image segmentation tasks into a unified framework, resulting in a informative outcome for autonomous driving scenarios. Our proposed network, Panoptic-DepthLab, builds upon a panoptic segmentation network with an encoder-decoder architecture and an ASPP module for capturing global context. We have further extended the network by incorporating an additional depth prediction branch. The fusion of panoptic segmentation and depth estimation results is achieved by averaging the pixel depths within each instance region, enabling comprehensive scene understanding. The effectiveness of our method has been evaluated on the Cityscape dataset, demonstrating high-quality segmentation and depth estimation results.

\bibliographystyle{IEEEbib}
\bibliography{reference}

\end{document}